\documentclass[11pt,journal]{IEEEtran}
\usepackage{blindtext}
\usepackage{graphicx}
\usepackage{cite}
\usepackage[justification=centering]{caption}
\usepackage{lettrine}
\usepackage{etoolbox}
\usepackage{amssymb}
\usepackage{mathtools}
\usepackage{lipsum}
\usepackage{tabularx}

\ifCLASSINFOpdf
\else
\fi

\begin{document}
\title{Face Recognition Machine Vision System Using Eigenfaces}

\author{Fares~Jalled,~\IEEEmembership{Moscow Institute of Physics \& Technology, Department of Radio Engineering \& Cybernetics~} 

\patchcmd 
\thanks{}
\thanks{\IEEEmembership{}}}

\markboth{}
{Shell \MakeLowercase{\textit{et al.}}: Bare Demo of IEEEtran.cls for Journals}

\maketitle

\begin{abstract}
Face Recognition is a common problem in Machine Learning. This technology has already been widely used in our lives. For example, Facebook can automatically tag people's faces in images, and also some mobile devices use face recognition to protect private security. Face images comes with different background, variant illumination, different facial expression and occlusion. There are a large number of approaches for the face recognition. Different approaches for face recognition have been experimented with specific databases which consist of single type, format and composition of image. Doing so, these approaches don't suit with different face databases. One of the basic face recognition techniques is eigenface which is quite simple, efficient, and yields generally good results in controlled circumstances. So, this paper presents an experimental performance comparison of face recognition using Principal Component Analysis (PCA) and Normalized Principal Component Analysis (N-PCA). The experiments are carried out on the ORL (ATT) and Indian face database (IFD) which contain variability in expression, pose, and facial details. The results obtained for the two methods have been compared by varying the number of training images.
MATLAB is used for implementing algorithms also.
\end{abstract}

\begin{IEEEkeywords}
\texttt{Face Recognition, Principal Component Analysis (PCA), Normalized Principal Component Analysis (N-PCA)}
\end{IEEEkeywords}

\section{Introduction}
\lettrine[lines=2]{F}
ace is a complex multidimensional structure and needs a good computing techniques for recognition.
Our approach treats face recognition as a two-dimensional recognition problem. Face recognition systems generally divided into two categories: identification or verification. In an identification set-up the similarity between a given face image and all the face images in a large database is computed, the top match is returned as the recognized identity of the subject. Different researchers for the face recognition system have proposed many linear and nonlinear statistical techniques. The PCA or Eigenfaces method is one of the most widely used linear statistical techniques reported by research community. In this paper, the N-PCA statistical technique is presented for the face recognition. The experimental results compare with the popular linear PCA statistical technique. The classification step chooses to be the simplest Euclidean distance classifier.
The rest of this paper is outlined as follows: section II illustrates the face recognition system. Section III explains some useful mathematical modeling. Section IV explains the Principal component analysis and the Normalized-PCA. Section V introduces the ORL and Indian face databases. The experimental analysis and results are discussed in Section VI. Finally, conclusion of the paper is given in Section VII.

\section{FACE RECOGNITION SYSTEM}
\begin{center}
\begin{figure}[!h]
\centering
\includegraphics[scale=0.5]{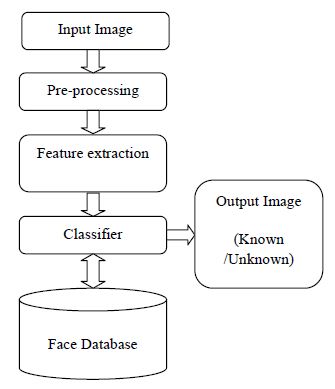}
\caption{Block Diagram of Face Recognition System}
\label{fig:block} 
\end{figure}
\end{center}

A pattern recognition task performed exclusively on faces is termed as face recognition. It can be described as classifying a face either known or unknown, after matching it with stored known individuals as a database. It is also advantageous to have a system that has the capability of learning to identify unknown faces. The outline of typical face recognition system implemented specifically for N-PCA is given in Figure \ref{fig:block}. There are five main functional blocks, whose responsibilities are as below.

\subsection{The Acquisition Module}
This is the entry point of the face recognition process. The user gives the face image as the input to face recognition system in this module.

\subsection{The Pre-Processing Module}
In this module the images are normalized to improve the recognition of the system. The pre-processing steps implemented are as follows:

\begin{itemize}
\item Image size normalization
\item Background removal
\item Translation and rotational normalizations
\item Illumination normalization
\end{itemize}

\subsection{The Feature Extraction Module}
After the pre-processing the normalized face image is given as input to the feature extraction module to find the key features that will be used for classification. The module composes a feature vector that is well enough to represent the face image.

\subsection{The Classification Module}
With the help of a pattern classifier, the extracted features of face image are compared with the ones stored in the face database. The face image is then classified as either known or unknown.

\subsection{Face Database}
It is used to match the test image with the train images stored in a database. If the face is recognized as “unknown”, face images can then be added to the database for further comparisons.

\begin{center}
\begin{figure}[!h]
\centering
\includegraphics[scale=0.4]{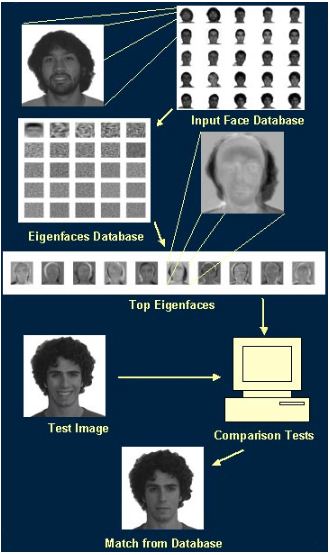}
\caption{Summary of Overall Face Recognition Process}
\label{fig:eigenface} 
\end{figure}
\end{center}

\section{Mathematical Modeling}
\subsection{\textbf{Eigen Values and Eigen Vectors}}
In linear algebra [12,13,14,15], the eigenvectors of a linear operator are non-zero vectors which, when operated by the operator, result in a scalar multiple of them. Scalar is then called Eigen value ($\lambda$)
 associated with the eigenvector (X). Eigen vector is a vector that is scaled by linear transformation. It is a property of matrix. When a matrix acts on it, only the vector magnitude is changed not the direction. 
 
\begin{equation}
AX = \lambda X, \qquad\text{where A is a vector function}\qquad
\end{equation}

\begin{equation}
(A - \lambda I)X = 0, \qquad\text{where I is the identity matrix}\qquad
\end{equation}

This is a homogeneous system of equations and form fundamental linear algebra. We know a non-trivial solution exists if and only if Det(A - $\lambda$I) = 0, where det denotes determinant.
When evaluated becomes a polynomial of degree n. This is called characteristic polynomial of A. If A is N by N then there are n solutions or n roots of the characteristic polynomial. Thus there are n Eigen values of A satisfying the equation.

\begin{equation}
AX_{i} = \lambda_iX_{i}, \qquad\text{where i = 1,2,3 ..,n}\qquad
\end{equation}

If the Eigen values are all distinct, there are n associated linearly independent eigen vectors, whose directions are unique, which span an n dimensional Euclidean space.

\subsection{\textbf{Face Image Representation}}
Training set of m images of size N x N are represented by vectors of size $N^{2}$. Each face is represented by $\Gamma_1, \Gamma_2, \Gamma_3, .., \Gamma_M.$
Feature vector of a face is stored in a N x N matrix. Now, this two dimensional vector is changed to one dimensional vector.

For example:
\[
  \begin{bmatrix}
    1 & 2  \\
    2 & 1 
  \end{bmatrix}
=
  \begin{bmatrix}
    1   \\
    2   \\
    2   \\
    1
  \end{bmatrix}
\]

Each face image is represented by the vector $\Gamma_i$

\[
\Gamma_1 = 
  \begin{bmatrix}
    1   \\
   -2   \\
    1   \\
   -3
  \end{bmatrix}
\Gamma_2 =
  \begin{bmatrix}
    1   \\
    3   \\
   -1   \\
    2
  \end{bmatrix} 
\Gamma_3 =
  \begin{bmatrix}
    2   \\
    1   \\
   -2   \\
    3
  \end{bmatrix} 
... \Gamma_M =
  \begin{bmatrix}
    1   \\
    2   \\
    2   \\
    1
  \end{bmatrix}      
\]

\subsection{\textbf{Mean Centered Images}}
Average face image is calculated by

\begin{equation}
\psi = \frac{1}{M} \sum_{i=1}^{M} \Gamma_i
\end{equation}

\[ 
  \begin{bmatrix}
    1   \\
   -2   \\
    1   \\
   -3
  \end{bmatrix}
+
  \begin{bmatrix}
    1   \\
    3   \\
   -1   \\
    2
  \end{bmatrix} 
+
  \begin{bmatrix}
    2   \\
    1   \\
   -2   \\
    3
  \end{bmatrix} 
+ ... +
  \begin{bmatrix}
    1   \\
    2   \\
    2   \\
    1
  \end{bmatrix}      
\rightarrow
  \begin{bmatrix}
   -1   \\
   -1   \\
    2   \\
   -3
  \end{bmatrix}
\]

\begin{equation}
\psi = \frac{(\Gamma_1+\Gamma_2+\Gamma_3+...+\Gamma_M)}{M}
\end{equation}

Each face differs from the average by $\Phi_i = \Gamma_i - \psi$ which is called mean centered image.

\[
\Phi_1 = 
  \begin{bmatrix}
    2   \\
   -1   \\
   -1   \\
    0
  \end{bmatrix}
\Phi_2 =
  \begin{bmatrix}
    2   \\
    4   \\
   -3   \\
    5
  \end{bmatrix} 
\Phi_3 =
  \begin{bmatrix}
    3   \\
    2   \\
   -4   \\
    6
  \end{bmatrix} 
... \Phi_M =
  \begin{bmatrix}
    2   \\
    3   \\
    0   \\
    4
  \end{bmatrix}      
\]

\subsection{\textbf{Covariance Matrix}}
A covariance matrix is constructed as $C = AA^{T}$

\begin{equation}
where \quad A = [\Phi_1,\Phi_2,\Phi_3,..,\Phi_M] \quad of \quad size \quad N^{2}x N^{2}
\end{equation} 

\[
A =
  \begin{bmatrix}
    2 & 3   \\
   -1 & -2  \\
   -1 & 1  \\
   0 & 2
  \end{bmatrix}
A^{T} =
  \begin{bmatrix}
    2 & -1 & -1 & 0 \\
   3 & -2 & 1 & 2
  \end{bmatrix}
\]

Size of covariance matrix will be $N^{2} x N^{2}$ 
(4 x 4 in this case). Eigen vectors corresponding to this covariance matrix is needed to be calculated, but that will be a tedious task therefore. For simplicity we calculate $A^{T}A$ which would be a 2 x 2 matrix in this case.

\[
A^{T}A =
\begin{bmatrix}
    6 & 7 \\
    7 & 18
  \end{bmatrix}
\]

Consider the eigenvectors $\nu_i$ of $A^{T}A$ such that

\begin{equation}
A^{T}AX_i = \lambda_iX_i
\end{equation} 

The eigenvectors $\nu_i$ of $A^{T}A$ are $X_{1}$ and $X_{2}$ which are 2 x 1. Now multiplying the above equation with A both sides we get

\begin{equation}
AA^{T}AX_i = A\lambda_iX_i
\end{equation} 

\begin{equation}
AA^{T}(AX_i) = \lambda_i(AX_i)
\end{equation} 

Eigen vectors corresponding to $AA^{T}$ can now be easily calculated with reduced dimensionality where $AX_{i}$ is the Eigen vector and $\lambda_i$ is the Eigen value.

\subsection{\textbf{Eigen Face Space}}
The Eigen vectors of the covariance matrix $AA^{T}$ are $AX^{i}$ which is denoted by $U^{i}$. $U^{i}$ resembles facial images which look ghostly and are called Eigen faces. Eigen vectors correspond to each Eigen face in the face space and discard the faces for which Eigen values are zero thus reducing the Eigen face space to an extent. The Eigen faces are ranked according to their usefulness in characterizing the variation among the images. A face image can be projected into this face space by

\begin{equation}
\Omega_k = U^{T}(\Gamma_k - \psi), \qquad\text{k = 1,..,M}\qquad 
\end{equation}

where: $(\Gamma_k\psi)$ is the mean centered image. Hence projection of each image can be obtained as $\Omega_1$ for projection of $image_{1}$ and $\Omega_2$ for projection of $image_{2}$ and hence forth.

\subsection{\textbf{Recognition Step}}
The test image, $\Gamma$, is projected into the face space to obtain a vector, $\Omega$ as $\Omega = U^{T}(\Gamma - \psi)$. The distance of $\Omega$ to each face is called Euclidean distance and defined by $\epsilon_k^{2} = ||\Omega - \Omega_k||^{2}$, k = 1,..,M where $\Omega_k$ is a vector describing the $k^{t}h$ face class. A face is classified as belonging to class k when the minimum $\epsilon_k$ is below some chosen threshold $\theta_c$, otherwise the face is classified as unknown. $\theta_c$ is half the largest distance between any two face images:

\begin{equation}
\theta_c = (\frac{1}{2})max_{j,k} ||\Omega_j - \Omega_k|| \qquad\text{j,k = 1,..,M}\qquad
\end{equation}

We have to find the distance $\epsilon$ between the original test image $\Gamma$ and its reconstructed image from the Eigen face $\Gamma_f$.

\begin{equation}
\epsilon^{2} = ||\Gamma - \Gamma^{f}||^{2} \qquad\text{where,}\qquad \Gamma^{f} = U*\Omega+\Psi
\end{equation}

if $\epsilon \geqslant \theta_c$ then input image is not even a face image and not recognized.\\
if $\epsilon < \theta_c$  and $\epsilon_k \geqslant \theta$ for all k then input image is a face image but it is recognized as unknown face.\\
if $\epsilon < \theta_c$ and  $\epsilon_k < \theta$  for all k then input images are the individual face image associated with the class vector $\Gamma_k$.

\section{Face Recognition Algorithm}
The face recognition system consists of two important steps, the feature extraction and the classification. Face recognition has a challenge to perform in real time. Raw face image may consume a long time to recognize since it suffers from a huge amount of pixels. One needs to reduce the amounts of pixels. This is called dimensionality reduction or feature extraction, to save time for the decision step. Feature extraction refers to transforming face space into a feature space. In the feature space, the face database is represented by a reduced number of features that retain most of the important information of the original faces. The most popular method to achieve this target is through applying the Eigenfaces algorithm. The Eigenfaces algorithm is a classical statistical method using the linear Karhumen-Loeve transformation (KLT) also known as Principal component analysis. The PCA calculates the eigenvectors of the covariance matrix of the input face space. These eigenvectors define a new face space where the images are represented. In contrast to linear PCA, N-PCA has been developed.

\subsection{\textbf{Principal Component Analysis}}
Principal component analysis (PCA) [1,2,3,4,5,6,10,11,13,14,15] is a statistical dimensionality reduction method, which produces the optimal linear least-square decomposition of a training set. In applications such as image compression and face recognition a helpful statistical technique called PCA is utilized and is a widespread technique for determining patterns in data of large dimension. PCA commonly referred to as the use of Eigen faces. The PCA approach is then applied to reduce the dimension of the data by means of data compression, and reveals the most effective low dimensional structure of facial patterns. The advantage of this reduction in dimensions is that it removes information that is not useful and specifically decomposes the structure of face into components which are uncorrelated and are known as Eigen faces. Each image of face may be stored in a 1D array which is the representation of the weighted sum (feature vector) of the Eigen faces. In case of this approach a complete front view of face is needed, or else the output of recognition will not be accurate. The major benefit of this method is that it can trim down the data required to recognize the entity to 1/1000th of the data existing. After feature extraction step next is the classification step which makes use of Euclidean Distance for comparing/matching of the test and trained images. In the testing phase each test image should be mean centred, now project the test image into the same eigenspace as defined during the training phase. This projected image is now compared with projected training image in eigenspace. Images are compared with similarity measures. The training image that is the closest to the test image will be matched and used to identify. Calculate relative Euclidean distance between the testing image and the reconstructed image of $i^{th}$ person, the minimum distance gives the best match.

\subsection{\textbf{Normalized Principal Component Analysis (N-PCA)}}
In contrast to linear PCA, N-PCA has been developed to give better results in terms of efficiency. N-PCA is an extension over linear PCA in which firstly normalization of images is done in order to remove the lightning variations and background effects and singular value decomposition (SVD) is used instead of eigen value decomposition (EVD), followed by the feature extraction steps of linear PCA and lastly in classification steps weights are calculated for matching purpose. The Normalized PCA is developed by taking into account the architecture described below: 

\begin{itemize}
\item[1] Collection of Images to make the Database. All the databases are collected and stored in a folder each containing images of 40 persons, having 10 images in different poses of each individual, making a total of 400 images in a folder for ORL Database.
\item[2] Checking whether Image is colored or gray: Initially the face image is taken and is checked whether it is colored or gray image by using the command size. If the face image taken is colored then it is converted to gray image to make it two dimensional (m x n). After this the class type of gray image is checked to make sure the image type is doubled or not. If it is not doubled than convert the image to double.
\item[3] Mean and Standard Deviation of Images is Calculated: After this the image is converted into a column vector of dimension mxn, so that a typical image of size 112x92 for example becomes a vector of dimension 10304, denoted as T. Now the mean of the image is calculated by using equation (4) and standard deviation (S) is calculated using equation (13).

\begin{equation}
S = \sqrt{\frac{1}{M} \sum_{i=1}^{M} \Gamma_i^{2}} 
\end{equation}

Also the standard deviation (ustd) and mean (um) is set practically approximately close to mean and standard deviation calculated above.

\item[4] Normalization: Next normalization of images is done using equation (14) in order to make all images of uniform dimension and to remove the effects due to background and illumination.

\begin{equation}
I = (T-\psi) * \frac{ustd}{S} + um 
\end{equation}

\item[5] Calculating Train Centred Images: Subtracting the mean from column vector matrix of training images in order to obtain the centred images.

\item[6] Calculating Eigen Vectors and Values (See Section III A.) 

\item[7] Creating Eigen faces: Eigen Vectors obtained after SVD are sorted in descending order and the top Eigen vectors are considered as Eigen Faces.

\item[8] Calculating Train Weights: Now the weights are determined by calculating the dot product of transpose of Eigen Faces matrix and Train Centred Images.

\item[9] Store Train Weights in Sink for Further Comparison: The above mentioned procedure from step 2) to 8) is applied on each train image and finally one by one train weights are stored in sink for further comparison.

\item[10] Euclidean distance Classifier: In this block firstly all steps from 2) to 7) are applied on the test image to compute its test weights and then the difference between this test weights and the train weights stored in sink is calculated using equation (15).

The Euclidean distance is defined by:

\begin{equation}
d(p_{j},q_{i}) = \sqrt{\sum_{j=1}^{k} (p_{j}-q_{ij})^{2}}
\end{equation}

where $p_{j}$ is the $j^{th}$ component of the test image vector, and $q_{ij}$ is the $j^{th}$ component of the train image vector $q_{i}$.

\item[11] Face Recognized: Finally the minimum distance gives the best match or can be said similar Face Identified. To ensure the result of face identification the Id of test image is compared with the Id of matched image. If both Ids are same then say face recognized otherwise face not recognized.

\end{itemize}

\section{Database of Faces} 
\subsection{Olivetti Research Laboratory (ORL)}
This paper considers the well known ORL face database that is taken at the Olivetti Research Laboratory in Cambridge, UK [7,8,11,13]. The ORL database contains 400 Grey images corresponding to ten different images of 40 distinct subjects.

\begin{center}
\begin{figure}[!h]
\centering
\includegraphics[scale=0.5]{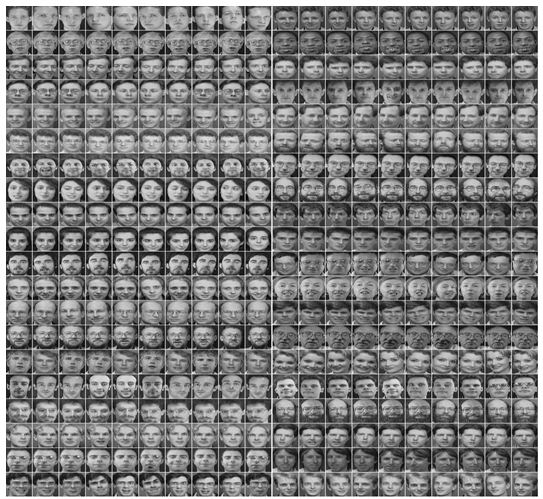}
\caption{Snapshot of the ORL Database}
\label{fig:database} 
\end{figure}
\end{center}

The images are taken at different times with different specifications; including varying slightly in illumination, different facial expressions i.e. open and closed eyes, smiling and non-smiling, and facial details i.e. glasses and no-glasses. All images were captured against a dark identical environment with the individuals in an upright, frontal position, as well as tolerance for some orientated and alternation of up to 20 degrees. There is some variation in scale of up to about 10\%. All the images are 8-bit gray scale with size 112 x 92 pixels. 




\subsection{Indian Face Database (IFD)}
The Indian Face Database contains human face images that are taken in the campus of Indian Institute of Technology Kanpur, in February, 2002 [9]. The IFD database contains images of 61 (39 males and 22 females) distinct subjects with eleven different poses for every entity. For some entities, a small number of extra images are also included if available. The background chosen for all the images is bright homogeneous and the individuals are in an upright, frontal position. For each individual, database included the following pose for the face: facing front, facing right, facing left, facing down, facing up, facing up towards right, facing up towards left. In addition to the discrepancy in pose, images with four emotions - neutral, smile, sad or disgust, laughter - are also incorporated for every entity. Some sample faces are shown in Figure 4.

\begin{center}
\begin{figure}[!h]
\centering
\includegraphics[scale=0.45]{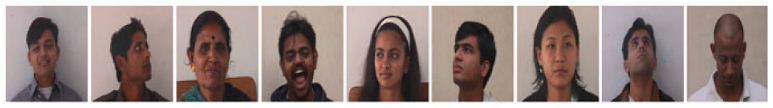}
\caption{Some Example Images from Indian Face Database}
\label{fig:indian} 
\end{figure}
\end{center}

\section{Results}
The experiments have been performed on ORL database with different number of training and testing images. Eigen faces are calculated by using PCA algorithm. The algorithm is developed in MATLAB. In the experimental set-up, the numbers of training images are varied from 80 percent to 40 percent that is initially 80\% of total images is used in training and remaining 20\% for testing and then the ratio was varied as 60/40 followed by 40/60. The experimental result shows that as the number of training images increases, efficiency of the system increases.

\begin{center}
\begin{figure}[!h]
\centering
\includegraphics[scale=0.6]{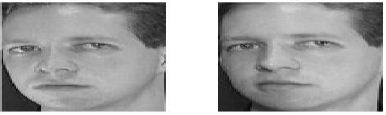}
\caption{Correct Face Recognition Result for ORL database}
\label{fig:correct} 
\end{figure}
\end{center}

\begin{center}
\begin{figure}[!h]
\centering
\includegraphics[scale=0.6]{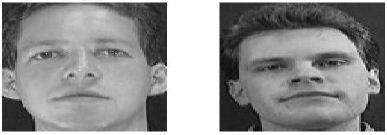}
\caption{Incorrect Face Recognition Result for ORL database}
\label{fig:incorrect} 
\end{figure}
\end{center}

\begin{center}
\begin{figure}[!h]
\centering
\includegraphics[scale=0.6]{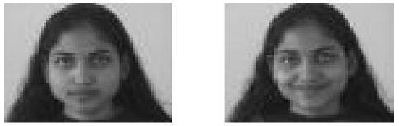}
\caption{Correct Face Recognition Result for IFD Database}
\label{fig:indiancorrect} 
\end{figure}
\end{center}

\begin{center}
\begin{figure}[!h]
\centering
\includegraphics[scale=0.6]{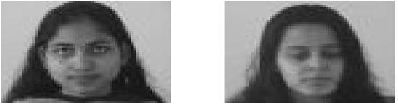}
\caption{Incorrect Face Recognition Result for IFD Database}
\label{fig:indianincorrect} 
\end{figure}
\end{center}

The accuracy of face recognition algorithm was measured by Euclidian distance between the test face and all train faces. The results of the experiments on IFD and ORL face database has been shown in next figures 5 to 8 respectively. For ORL database figure 5 shows the correct face recognition result using N-PCA method and figure 6 illustrates incorrect recognition result. Similarly for IFD, using N-PCA method figure \ref{fig:indiancorrect} shows the correct face recognition result and figure \ref{fig:indianincorrect} presents incorrect recognition result.

\begin{center}
\begin{figure}[!h]
\centering
\includegraphics[scale=0.6]{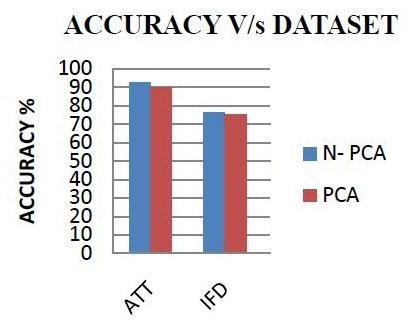}
\caption{Accuracy v/s Dataset for 80/20 Ratio}
\label{fig:1} 
\end{figure}
\end{center}

\begin{center}
\begin{figure}[!h]
\centering
\includegraphics[scale=0.6]{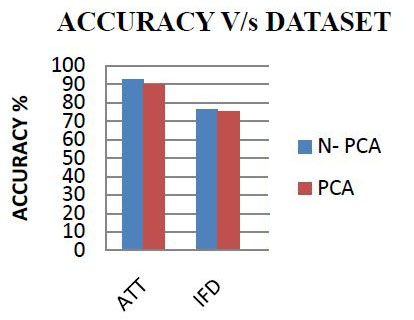}
\caption{Accuracy v/s Dataset for 60/40 Ratio}
\label{fig:2} 
\end{figure}
\end{center}

\begin{center}
\begin{figure}[!h]
\centering
\includegraphics[scale=0.6]{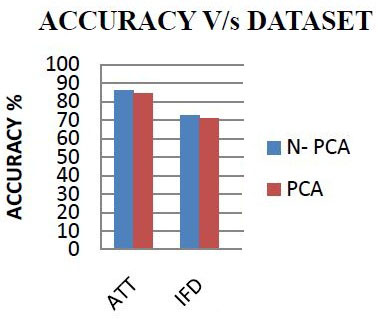}
\caption{Accuracy v/s Dataset for 40/60 Ratio}
\label{fig:3} 
\end{figure}
\end{center}

The performance of the proposed feature extraction scheme (N-PCA) is compared with PCA in terms of percentage of recognition by varying the database ratio in training and testing and the results are presented in table 1 for ORL and IFD databases. The accuracy v/s dataset graphs are plotted for different training and testing database ratio in figure \ref{fig:1}, \ref{fig:2} and \ref{fig:3}.

\vspace*{0.5cm}

\begin{center}
\begin{tabular}{|l|l|r|r|}
\hline
Train/TPD & DATASET & ORL & IFD \\
\hline
80/20 & N-PCA & 93.75 & 76.67 \\
\cline{2-4} 
& PCA & 92.50 & 74.17 \\
\hline
60/40 & N-PCA & 92.50 & 76.25 \\
\cline{2-4} 
& PCA & 90.00 & 75.00 \\
\hline
40/60 & N-PCA & 85.83 & 72.53 \\
\cline{2-4} 
& PCA & 85.42 & 71.98 \\
\hline
\end{tabular}
\end{center}

\vspace*{0.5cm}

Tab. 1 Accuracy for N-PCA compared with PCA on
different number of training and testing images.

*TPD: Test Percentage of Database.

\section{Conclusion}
The Eigenface approach for Face Recognition process is fast and simple which works well under constrained environment. It is one of the best practical solutions for the problem of face recognition. Instead of searching large database of faces, it is better to give small set of likely matches. By using Eigenface approach, this small set of likely matches for given images can be easily obtained. The face recognition system consists of two important steps, the feature extraction and the classification. This paper investigates the N-PCA function improvement over the principal component analysis (PCA) for feature extraction. The experiments carried out to investigate the performance of N-PCA by comparing it with the performance of the PCA. The analysis is done with ORL and IFD dataset, the recognition performance for both algorithms are evaluated and the experimental results shows that N-PCA gives a better recognition rate. It is clear that the comparison shows that accuracy for PCA is 92.50\% and 74.17\% , while for N-PCA is 93.75\% and 76.67\% on ORL and IFD for 80/20 Ratio. The promising results indicate that N-PCA can emerge as an effective solution to face recognition  problems in the future. This can give us much better results if combined with other techniques.

\end{document}